\pdfoutput=1

\documentclass[11pt]{article}

\usepackage[]{ACL2023}

\usepackage{times}
\usepackage{latexsym}

\usepackage[T1]{fontenc}

\usepackage[utf8]{inputenc}

\usepackage{microtype}
\usepackage{inconsolata}
\usepackage{algorithm}
\usepackage{algorithmic}
\usepackage{listings}
\usepackage{newfloat}
\usepackage{listings}
\usepackage{titlesec}
\usepackage{verbatim}
\usepackage{multirow}
\usepackage{graphicx}
\usepackage{booktabs}
\usepackage{amssymb}
\usepackage{amsmath}
\usepackage{bm}
\usepackage{color, soul}
\usepackage{makecell}
\usepackage[capitalise]{cleveref}

\usepackage[algo2e]{algorithm2e} 

\definecolor{wingreen}{rgb}{0,0.45,0.24}
\definecolor{losered}{rgb}{1.0,0.1,0.24}

\newcommand{\methodname}[1]{DP}
\newcommand{\basea}[1]{base}
\newcommand{\baseb}[1]{base}
\newcommand{\data}[1]{data}

\title{Exploring the Limits of ChatGPT for Query or Aspect-based Text Summarization}

\author{Xianjun Yang$^{1}$ \qquad Yan Li$^{2}$ \qquad Xinlu Zhang$^{1}$ \qquad Haifeng Chen$^{3}$ \qquad Wei Cheng$^{3}$\\
$^{1}$ University of California, Santa Barbara \qquad $^{2}$ Microsoft \\ \qquad $^{3}$ NEC Laboratories America } 

\begin{document}
\maketitle

\begin{abstract}

Text summarization has been a crucial problem in natural language processing (NLP) for several decades. It aims to condense lengthy documents into shorter versions while retaining the most critical information. Various methods have been proposed for text summarization, including extractive and abstractive summarization. The emergence of large language models (LLMs) like GPT3 and ChatGPT has recently created significant interest in using these models for text summarization tasks. Recent studies \cite{goyal2022news, zhang2023benchmarking} have shown that LLMs-generated news summaries are already on par with humans. However, the performance of LLMs for more practical applications like aspect or query-based summaries is underexplored. To fill this gap, we conducted an evaluation of ChatGPT's performance on four widely used benchmark datasets, encompassing diverse summaries from Reddit posts, news articles, dialogue meetings, and stories. Our experiments reveal that ChatGPT's performance is comparable to traditional fine-tuning methods in terms of Rouge scores. Moreover, we highlight some unique differences between ChatGPT-generated summaries and human references, providing valuable insights into the superpower of ChatGPT for diverse text summarization tasks. Our findings call for new directions in this area, and we plan to conduct further research to systematically examine the characteristics of ChatGPT-generated summaries through extensive human evaluation.

\end{abstract}

\section{Introduction}

Text summarization has long been a pivotal challenge in the field of Natural Language Processing (NLP). The main objective of this task is to succinctly condense a lengthy document into a shorter version while ensuring that the most crucial information is preserved. With the recent rise of advanced language models like ChatGPT, there has been a heightened interest in leveraging these models for text summarization tasks. However, it is noteworthy that the majority of existing research studies \cite{goyal2022news, zhang2023benchmarking} have primarily concentrated on generating a general summary for news-related content.

Aspect- or query-based summarization represents a more diverse and nuanced form of text summarization that has garnered significant attention within the NLP community. Unlike generic summarization, these tasks involve generating summaries that are customized to particular aspects or queries, rather than a single condensed version of the entire document. Consequently, this approach demands a deeper level of comprehension of the document, with respect to the specific interests and needs of the users.

In this paper, we present a comprehensive evaluation of ChatGPT's performance on four distinct aspect-based and query-based text summarization tasks. Our experimental analysis indicates that ChatGPT's summarization capabilities are on par with traditional fine-tuning methods, based on Rouge scores. The outcomes of this study offer valuable perspectives on the potential of ChatGPT for text summarization tasks, and emphasize the necessity for innovative approaches in this field. The achievement of ChatGPT in text summarization tasks holds promising implications for the development of practical and effective summarization systems.

Recently, a study by \cite{goyal2022news} demonstrated that, while GPT-3 generated summarizations achieved lower Rouge scores compared to traditional fine-tuning methods, human annotators favored the text generated by GPT-3. In addition, a thorough analysis of large language models for news summarization by \cite{zhang2023benchmarking} concluded that the summarizations produced by these models were already comparable to those generated by humans, which they attributed to instruction tuning. Notably, \cite{Bang2023AMM} conducted a comprehensive investigation of ChatGPT's multi-task, multilingual, and multimodal evaluation, including text summarization as a case study, and arrived at similar conclusions. However, the evaluation datasets used in the News domain for general summarization were not explicitly designed for text summarization and focused solely on general summarization. As such, we aim to explore how ChatGPT performs in the diverse summarization of lengthy articles across multiple domains, using high-quality data.

This work makes several significant contributions, including:

\begin{itemize}
\item Being the first systematic attempt to extend the usage of LLMs beyond generic summarization and examining the performance of ChatGPT in aspect or query-based summarization.
\item Demonstrating that ChatGPT-generated diverse specific summaries are highly comparable to traditional fine-tuning methods in terms of Rouge scores.
\item Conducting an in-depth analysis of the LLM-generated summaries and identifying several potential future research directions that could leverage the strengths of LLMs.
\end{itemize}

Together, these contributions provide novel insights into the capabilities of ChatGPT for diverse text summarization tasks and underscore the potential of LLMs as a powerful tool for NLP research.

\section{Related Work}
\subsection{Aspect or Query-based Summarization}
Aspect- and query-based summarization are two critical forms of text summarization that differ significantly from general summaries in that they are not input-agnostic. These tasks aim to generate summaries that are tailored to specific aspects or queries for various types of content, such as news articles \cite{kulkarni2020aquamuse}, meetings \cite{zhong-etal-2021-qmsum}, stories \cite{wang2022squality}, and Wikipedia articles \cite{yang2022oasum}. This approach contrasts with previous methods such as CNN/DM \cite{hermann2015teaching} or XSUM \cite{narayan2018don}, which focus on developing a single generic summary of the entire document. By leveraging aspect- or query-based summarization, it is possible to create more targeted and personalized summaries that cater to the specific interests and needs of different users.

Aspect-based and query-based summarization can be accomplished using a variety of methods, including end-to-end and extract-then-summarize approaches. End-to-end summarization directly produces the summaries without manipulating the original inputs. In contrast, the extract-then-summarize approach involves identifying and extracting the most important sentences or phrases from the original document to form a shorter document, which is then summarized to fit the input limit of language models, such as BART, which has a token limit of 1024 \cite{lewis-etal-2020-bart}. Additionally, abstractive summarization methods aim to generate new sentences or phrases that summarize the original document's content, rather than simply extracting and rephrasing existing text. There is no one-size-fits-all approach to aspect- and query-based summarization, and the choice of method depends on factors such as the size and complexity of the input, the desired length and level of detail of the summary, and the target audience's needs and preferences.

\subsection{Large Language Models}
In recent years, large language models such as GPT-3 \cite{brown2020language} and ChatGPT have garnered substantial interest in the field of natural language processing. These models are trained on vast quantities of text data and have achieved remarkable performance on a range of NLP tasks, including text classification, question answering, and machine translation.

Several studies have investigated using large language models for text summarization tasks. For instance, Goyal et al. \cite{goyal2022news} observed that while GPT-3-generated summaries obtained slightly lower Rouge scores than traditional fine-tuning methods, human evaluators preferred the former. Similarly, Zhang et al. \cite{zhang2023benchmarking} reported that LMM-generated summaries were considered as good as human-written summaries in the News domain. Besides, \cite{Qin2023IsCA} also competitively examine the performance of ChatGPT and GPT-3.5 for various tasks, including dialogue summarization dataset SAMSUm \cite{gliwa-etal-2019-samsum}

As recent studies have highlighted the potential of large language models for text summarization, it is essential to further investigate their performance on diverse summarization tasks in various domains. Our work aims to contribute to this ongoing research by evaluating the capabilities of ChatGPT on aspect-based and query-based summarization tasks and providing insights into its strengths and limitations.

\section{Task Formulation}
\begin{table*}[t]
    \centering
    \scalebox{0.85}{
    \begin{tabular}{c|c|c|c|c|c }
        \toprule
        \textbf{Type} & \textbf{Dataset} & \textbf{Domain} & \#\textbf{Input Tk.} &\#\textbf{Output Tk.} & \#\textbf{Asp. Type}  \\
        \hline
        \multirow{2}{*}{Query}
        & QMSum \cite{zhong-etal-2021-qmsum} & Meeting & 9,070(2,505*) & 70 & 1,566  \\
        & SQuALITY \cite{wang2022squality} & Story & 6,052 & 252 & 437   \\
        \hline
        \multirow{2}{*}{Aspect}
        & CovidET \cite{zhan2022you}  & Reddit &  192 & 27 & 7  \\
        & NEWTS \cite{bahrainian-etal-2022-newts}  & News & 602 & 74 & 50 \\

        \bottomrule
    \end{tabular}
    }
    \caption{
    Statistics of the query/aspect-based summarization datasets that we used.
    \#\textbf{Input Tk.} and \#\textbf{Output Tk.} represent the number of input and output token lengths, respectively. \#\textbf{Asp. Type} is the number of all aspect types. 2,505* represents the average token number in golden inputs.
    }
    \label{tab:big_table}
\end{table*}

\begin{table*}[t]
\centering
\scalebox{1.0}{
\begin{tabular}{l@{\hspace{0.8\tabcolsep}}|c|cccc}
\hline
{\textbf{Datasets}} & Models & R-1  & R-2 & R-L & R-Lsum  \\
\hline
\multirow{2}{*}{\textit{CovidET}} & \textit{Fine-tuning} & $\textbf{26.19}$ &$\textbf{6.85}$ & $\textbf{17.86}$ & $\textbf{20.82}$ \\
& \textit{ChatGPT}  & $20.81 $ & $3.99 $ & $15.35$ & $15.36$ \\
\hline
\multirow{2}{*}{\textit{NEWTS}} & \textit{Fine-tuning} & $31.78$ & $10.83$ & $20.54$ & $-$ \\
& \textit{ChatGPT} & $\textbf{32.54}$ & $\textbf{11.37} $ & $\textbf{20.74}$ & $20.74 $ \\
\hline
\multirow{2}{*}{\textit{QMSum}} & \textit{Fine-tuning} & $\textbf{32.29}$ & $8.67$ & $\textbf{28.17}$ & $-$\\
& \textit{ChatGPT} & $28.34$ & $\textbf{8.74}$ & $17.81 $ & $18.01$\\
\hline
\multirow{2}{*}{\textit{QMSum(Golden)}} & \textit{Fine-tuning} & $36.06$ & $11.36$ & $\textbf{31.27}$ & $-$\\
& \textit{ChatGPT} & $\textbf{36.83} $ & $\textbf{12.78}$ & $24.23$ & $ 24.19 $   \\
\hline
\multirow{2}{*}{\textit{SQuaLITY}} & \textit{Fine-tuning} & $\textbf{38.20}$ & $\textbf{9.00}$ & $\textbf{20.20}$ & $-$ \\
& \textit{ChatGPT} & $37.02 $ & $8.19$ & $18.45$ & $22.56$ \\
\hline
 \multirow{2}{*}{\textit{Avg.}} & \textit{Fine-tuning} & $\textbf{32.90}$ & $\textbf{9.34}$ & $\textbf{23.61}$ & $-$ \\
& \textit{ChatGPT} & $30.94$ & $8.96 $ & $19.22 $ & $-$  \\
\hline
\end{tabular}
}
\caption{ Comparison between ChatGPT zero-shot performance with previous Fine-tuning(FT) results.}
\label{tab:res}
\end{table*}

\begin{table*}[t]
\centering
\scalebox{0.85}{
\begin{tabular}{l@{\hspace{0.8\tabcolsep}}|c|ccccccc}
\hline
{\textbf{Datasets}} & Text & U-1-gram  & U-2-gram & U-3-gram & U-4-gram & Coverage & Density & Compression \\
\hline 
\multirow{2}{*}{\textit{CovidET}} & \textit{Reference} & $0.59$ & $0.95$ & $0.99$ & $0.99$ & $0.60$ & $0.90$ & $11.84$ \\
& \textit{ChatGPT}  & $0.42$ & $0.86$ & $0.97$ & $0.99$ & $0.66$ & $1.19$ & $6.92$ \\
\hline
\multirow{2}{*}{\textit{NEWTS}} & \textit{Reference} & $0.30$ & $0.69$ & $0.84$ & $0.90$ & $0.78$ & $2.84$ & $9.66$ \\
& \textit{ChatGPT} & $0.15$ & $0.50$ & $0.69$ & $0.81$ & $0.89$ & $4.13$ & $4.03$ \\
\hline
\multirow{2}{*}{\textit{QMSum}} & \textit{Reference} & $0.19$ & $0.65$ & $0.88$ & $0.96$ & $0.87$ & $2.15$ & $99.76$ \\
& \textit{ChatGPT} & $0.26$ & $0.73$ & $0.89$ & $0.95$ & $0.84$ & $1.97$ & $44.86$ \\
\hline
\multirow{2}{*}{\textit{QMSum(Golden)}} & \textit{Reference} & $0.29$ & $0.69$ & $0.88$ & $0.95$ & $0.77$ & $2.05$ & $15.68$ \\
& \textit{ChatGPT} & $0.26$ & $0.66$ & $0.85$ & $0.93$ & $0.79$ & $2.15$ & $10.74$ \\
\hline
\multirow{2}{*}{\textit{SQuaLITY}} & \textit{Reference} & $0.25$ &$0.85$ & $0.97$ & $0.99$ & $0.83$ & $1.55$ & $32.57$ \\
& \textit{ChatGPT} & $0.33$ & $0.82$ & $0.93$ & $0.97$ & $0.81$ & $1.81$ & $24.83$ \\
\hline

\end{tabular}
}
\caption{ Comparison between ChatGPT zero-shot performance with references on various metrics. U-1/2/3/4-gram represents unique 1/2/3/4 gram.}
\label{tab:res_stats}
\end{table*}

Aspect- and query-based summarization are essential tasks for text summarization because they are considered more challenging and valuable for real-world production. These tasks aim to generate a summary tailored to specific aspects or queries, rather than generating a single generic summary of the entire document.

In this study, we evaluated the performance of ChatGPT on a series of aspect- and query-based text summarization benchmarks. The main steps of our experimental methodology are as follows:

\textbf{Data collection:} We selected publicly available datasets as listed in Tabel \ref{tab:big_table}, ensuring that they are consistent with previous finetuning methods.

\textbf{Model evaluation:} We conducted an evaluation of ChatGPT's performance on question and answer pairs, utilizing Rouge scores as our evaluation metric. Due to the lack of an API provided by ChatGPT for processing large amounts of input data, we manually evaluated 100 examples selected at random from each test set on the ChatGPT platform. Previous research efforts \cite{goyal2022news, zhang2023benchmarking} have also been limited in their testing of GPT-3 on a small number of instances.

\subsection{Prompts}\label{sec:prompt}
Here we list prompts used in our experiments for generated summaries.

\textbf{SQuALITY} The prompt is \textit{Q: Query. Answer the question in around 200 words. Article: story.} for a specific question, while \textit{Q: Query. Answer the question in around 450 words. Article: story.} for a general question. In the second case, if the generated summaries are much shorter than 450 words, we will specify \textit{Your response is too short. Please answer it in around 450 words.} to get a second-round conversation.\

\textbf{QMSum} The prompt is \textit{Q: Query. Article: meeting} or \textit{Q: Query. Article: golden meeting}, where \textit{meeting} is the initial meeting, while \textit{golden meeting} is the provided golden spans of sentences in the original long meeting.

\textbf{CovidET} The prompt is \textit{Q: Summarize this article with respect to Aspect within one short sentence. Article0. A: Answer0. Q: Summarize this article with respect to Aspect within one short sentence. Article. A: }, where \textit{Article0} and \textit{Answer0} are randomly picked from the training instances to serve as the in-context one-shot example. When \textit{within one short sentence} is omitted, we observed the summaries are much longer and even on par with the input article. We observe a significantly lower performance for zero-shot from preliminary experiments, likely due to the concise input and output. Thus we adopt 1-shot experiments for CovidET.

\textbf{NEWTS} The prompt is \textit{Article. Summarize this article with respect to Aspect: }, where the \textit{Aspect} is some continuous words serving as certain topics.

Unless expressly stated, we did not conduct any further conversation with ChatGPT to correct the answer. We test zero-shot performance for all datasets except for CovidET.

\section{Experiments and Analysis}
\subsection{Experiments.} We use the ChatGPT \footnote{https://chat.openai.com/chat} platform for conducting our experiments between February 10 to Feb. 15, 2022. To eliminate the effects of historical chats, we clear each conversation after generating each summary. For all datasets, we use their originally released corpus as testing examples.

\subsection{Analysis} The overall results are shown in Table \ref{tab:res}. As we can see, ChatGPT achieves comparable performance with traditional finetuning methods in all datasets. Surprisingly, when provided with golden annotation of meeting spans in QMSum, ChatGPT even outperforms finetuning in terms  of Rouge-1 and Rouge-2, though Rouge-L lags. The worst performance of ChatGPT is observed in CovidET, where the inputs are usually around 128 words, and the summary is almost always comprised of only one sentence with around 20 words. We attribute this low performance to the untypical length in CovidET, compared with most summarization datasets where the inputs and outputs are always much longer.
Regarding the news domain, ChatGPT outperforms finetuning in terms of all Rouge scores. This finding is consistent with the previous conclusion that Instruct-GPT could achieve near SOTA performance for a general summary in the news domain. We suspect this is due to the large availability of news corpus for pre-training.

In the context of meeting dialogue summarization in QMSum, we examine two scenarios where the input length exceeds the maximum token limit of ChatGPT. In the first scenario, we extract and summarize the input by splitting it into two parts, asking ChatGPT to extract salient information about the question, and then combining the extracted parts and performing a second round of summarization. We use finetuning as the comparing baseline for the extraction and summarization. The results demonstrate that ChatGPT outperforms finetuning in terms of Rouge-2 and exhibits comparable performance on Rouge-1, but performs significantly worse on Rouge-L. However, when given golden spans, ChatGPT performs slightly better on Rouge-1 and Rouge-2 in the zero-shot setting than fine-tuning on the golden inputs.  In both cases, we observe a significant gap in Rouge-L, likely caused by the data feature of oral dialogues. Rouge-L stands for Longest Common Subsequence (LCS), and the finetuning could bias toward the close-to-oral dialogues summaries, but ChatGPT tends to make more formal summaries. As a result, lower Rouge-L does not necessarily imply worse performance, and we intend to evaluate it using human evaluations in the future.

Lastly, since the inputs in the English story summaries dataset SQuaLITY are usually longer than 3000 words, we directly truncate them to fit into ChatGPT, following finetuning baseline that also truncates them. Notice that for some queries that can not be answered by the truncated paragraphs, ChatGPT will directly return the message that it could not be answered. We abandon such instances since the summaries are meaningless. We also use two prompts for a general summary of the story's plot or some specific questions, observing that they correspond to very different summaries as detailed in \ref{sec:prompt}. From the results, we again see similar performance on all evaluations of Rouge scores, with ChatGPT lagging by only 1 point. 

Following \cite{grusky-etal-2018-newsroom}, we use the Coverage, Density, and Compression to measure to what extent the summary is derivative of a text, how well the word sequence of a summary can be described as a series of extractions, and word ratio between the article and summary. We also use unique n-grams(n=$1, 2, 3, 4$) to denote how many unique words are presented in the summaries. The results are calculated for golden references and ChatGPT-generated summaries in Table \ref{tab:res_stats}. As we can see, ChatGPT-generated text consistently achieves a lower compression ratio, indicating that it prefers generating more extended summaries. While for coverage and density, there is no apparent difference for all scenarios.
For articles with long inputs like QMSum and SQuaLITY, the unique n-grams(n=$1,2$) are usually higher for ChatGPT while lower for unique n-grams(n=$3,4$), suggesting that ChatGPT-summaries are more abstractive in terms of short words. While for the remaining datasets, ChatGPT almost always generates less fraction of unique n-grams.

\subsection{Insights}
In the above, we see the exceptional summarization ability of ChatGPT across Reddit posts, news, dialogue, and meeting domains toward various aspects and queries. From some case studies as shown in Table \ref{tab:case} and \ref{tab:case2} in the Appendix, we can tell the ChatGPT-generated summaries are surprisingly good and even better than the given references. We leave the complete human evaluation for future work.
Considering the zero-shot performance of ChatGPT does not involve any additional labeling efforts for training data, and ChatGPT could even achieve better results given more appropriate prompts or multiple conversations for self-correction, we believe it is time for rethinking future directions for various text summarization tasks. Theoretically speaking, our experiments merely establish the lower threshold of ChatGPT's capabilities for aspect or query-based summarization. We are of the conviction that in the near future (possibly within a few months), ChatGPT could conceivably exceed the performance achieved through fine-tuning, owing to the utilization of superior prompts, the incorporation of multiple conversations involving self-correction, and the self-enhancement of ChatGPT itself.

\section{Conclusion}

In this paper, we evaluated the performance of ChatGPT on aspect- and query-based text summarization tasks across diverse domains. These results demonstrate the super ability of ChatGPT for various controllable text summarization tasks. However, the Rouge score might not be a good indicator for evaluating the performance of ChatGPT in text summarization tasks. We will conduct human evaluations of the generated text shortly to provide a more comprehensive assessment of ChatGPT's performance. In conclusion, our findings suggest that ChatGPT holds promise as a powerful tool for text summarization and lays the insights for future research in this area.

In the era of ChatGPT, we conclude some future directions that are worthy of investigating: \\
\textbf{1. Retrieval module:} In view of the fact that the training of large language models such as ChatGPT is constantly challenged by constraints on input length, the solution lies in the adoption of a lighter model such as LED \cite{beltagy2020longformer}, which is adept at swiftly retrieving significant sentences from lengthy inputs. By integrating LED, ChatGPT can effectively tackle the processing of lengthy documents. \\
\textbf{2. GPT-generated text detection:} Although ChatGPT-generated summaries are already very fluent and consistent, they might still include nonfactual or biased summaries. Thus it is important to develop tools for detecting such ChatGPT-generated summaries before they are widely deployed for real applications.\\
\textbf{3. Better prompts:} Given that our preliminary experiments have not thoroughly explored the possibility space of prompts, and we have yet to examine multiple conversations to refine our summaries, we are of the opinion that improving summary quality through enhanced prompting would be a topic of independent interest.

\section{Limitations}
In this study, we evaluated the performance of ChatGPT on aspect- and query-based text summarization tasks. However, our experiments were limited by the maximum input sequence length of ChatGPT, which is currently set to around 5000 tokens. This limitation could impact the generalizability of our results to other text summarization tasks and datasets, as the length of documents can vary widely in real-world applications. 

It is important to note that the results of this study should not be used to make decisions that could negatively impact individuals or groups. Further research is needed to thoroughly assess the ethical implications of using language models such as ChatGPT for text summarization tasks, particularly regarding fairness, bias, and factuality.

\bibliography{anthology,custom}

\begin{thebibliography}{17}
\expandafter\ifx\csname natexlab\endcsname\relax\def\natexlab#1{#1}\fi

\bibitem[{Bahrainian et~al.(2022)Bahrainian, Feucht, and
  Eickhoff}]{bahrainian-etal-2022-newts}
Seyed~Ali Bahrainian, Sheridan Feucht, and Carsten Eickhoff. 2022.
\newblock \href {https://doi.org/10.18653/v1/2022.findings-acl.42} {{NEWTS}: A
  corpus for news topic-focused summarization}.
\newblock In \emph{Findings of the Association for Computational Linguistics:
  ACL 2022}, pages 493--503, Dublin, Ireland. Association for Computational
  Linguistics.

\bibitem[{Bang et~al.(2023)Bang, Cahyawijaya, Lee, Dai, Su, Wilie, Lovenia, Ji,
  Yu, Chung, Do, Xu, and Fung}]{Bang2023AMM}
Yejin Bang, Samuel Cahyawijaya, Nayeon Lee, Wenliang Dai, Dan Su, Bryan Wilie,
  Holy Lovenia, Ziwei Ji, Tiezheng Yu, Willy Chung, Quyet~V. Do, Yan Xu, and
  Pascale Fung. 2023.
\newblock A multitask, multilingual, multimodal evaluation of chatgpt on
  reasoning, hallucination, and interactivity.

\bibitem[{Beltagy et~al.(2020)Beltagy, Peters, and
  Cohan}]{beltagy2020longformer}
Iz~Beltagy, Matthew~E Peters, and Arman Cohan. 2020.
\newblock Longformer: The long-document transformer.
\newblock \emph{arXiv preprint arXiv:2004.05150}.

\bibitem[{Brown et~al.(2020)Brown, Mann, Ryder, Subbiah, Kaplan, Dhariwal,
  Neelakantan, Shyam, Sastry, Askell et~al.}]{brown2020language}
Tom Brown, Benjamin Mann, Nick Ryder, Melanie Subbiah, Jared~D Kaplan, Prafulla
  Dhariwal, Arvind Neelakantan, Pranav Shyam, Girish Sastry, Amanda Askell,
  et~al. 2020.
\newblock Language models are few-shot learners.
\newblock \emph{Advances in neural information processing systems},
  33:1877--1901.

\bibitem[{Gliwa et~al.(2019)Gliwa, Mochol, Biesek, and
  Wawer}]{gliwa-etal-2019-samsum}
Bogdan Gliwa, Iwona Mochol, Maciej Biesek, and Aleksander Wawer. 2019.
\newblock \href {https://doi.org/10.18653/v1/D19-5409} {{SAMS}um corpus: A
  human-annotated dialogue dataset for abstractive summarization}.
\newblock In \emph{Proceedings of the 2nd Workshop on New Frontiers in
  Summarization}, pages 70--79, Hong Kong, China. Association for Computational
  Linguistics.

\bibitem[{Goyal et~al.(2022)Goyal, Li, and Durrett}]{goyal2022news}
Tanya Goyal, Junyi~Jessy Li, and Greg Durrett. 2022.
\newblock News summarization and evaluation in the era of gpt-3.
\newblock \emph{arXiv preprint arXiv:2209.12356}.

\bibitem[{Grusky et~al.(2018)Grusky, Naaman, and
  Artzi}]{grusky-etal-2018-newsroom}
Max Grusky, Mor Naaman, and Yoav Artzi. 2018.
\newblock \href {https://doi.org/10.18653/v1/N18-1065} {{N}ewsroom: A dataset
  of 1.3 million summaries with diverse extractive strategies}.
\newblock In \emph{Proceedings of the 2018 Conference of the North {A}merican
  Chapter of the Association for Computational Linguistics: Human Language
  Technologies, Volume 1 (Long Papers)}, pages 708--719, New Orleans,
  Louisiana. Association for Computational Linguistics.

\bibitem[{Hermann et~al.(2015)Hermann, Kocisky, Grefenstette, Espeholt, Kay,
  Suleyman, and Blunsom}]{hermann2015teaching}
Karl~Moritz Hermann, Tomas Kocisky, Edward Grefenstette, Lasse Espeholt, Will
  Kay, Mustafa Suleyman, and Phil Blunsom. 2015.
\newblock Teaching machines to read and comprehend.
\newblock \emph{Advances in neural information processing systems}, 28.

\bibitem[{Kulkarni et~al.(2020)Kulkarni, Chammas, Zhu, Sha, and
  Ie}]{kulkarni2020aquamuse}
Sayali Kulkarni, Sheide Chammas, Wan Zhu, Fei Sha, and Eugene Ie. 2020.
\newblock Aquamuse: Automatically generating datasets for query-based
  multi-document summarization.
\newblock \emph{arXiv preprint arXiv:2010.12694}.

\bibitem[{Lewis et~al.(2020)Lewis, Liu, Goyal, Ghazvininejad, Mohamed, Levy,
  Stoyanov, and Zettlemoyer}]{lewis-etal-2020-bart}
Mike Lewis, Yinhan Liu, Naman Goyal, Marjan Ghazvininejad, Abdelrahman Mohamed,
  Omer Levy, Veselin Stoyanov, and Luke Zettlemoyer. 2020.
\newblock \href {https://doi.org/10.18653/v1/2020.acl-main.703} {{BART}:
  Denoising sequence-to-sequence pre-training for natural language generation,
  translation, and comprehension}.
\newblock In \emph{Proceedings of the 58th Annual Meeting of the Association
  for Computational Linguistics}, pages 7871--7880, Online. Association for
  Computational Linguistics.

\bibitem[{Narayan et~al.(2018)Narayan, Cohen, and Lapata}]{narayan2018don}
Shashi Narayan, Shay~B Cohen, and Mirella Lapata. 2018.
\newblock Don't give me the details, just the summary! topic-aware
  convolutional neural networks for extreme summarization.
\newblock \emph{arXiv preprint arXiv:1808.08745}.

\bibitem[{Qin et~al.(2023)Qin, Zhang, Zhang, Chen, Yasunaga, and
  Yang}]{Qin2023IsCA}
Chengwei Qin, Aston Zhang, Zhuosheng Zhang, Jiaao Chen, Michihiro Yasunaga, and
  Diyi Yang. 2023.
\newblock Is chatgpt a general-purpose natural language processing task solver?

\bibitem[{Wang et~al.(2022)Wang, Pang, Chen, Phang, and
  Bowman}]{wang2022squality}
Alex Wang, Richard~Yuanzhe Pang, Angelica Chen, Jason Phang, and Samuel~R
  Bowman. 2022.
\newblock Squality: Building a long-document summarization dataset the hard
  way.
\newblock \emph{arXiv preprint arXiv:2205.11465}.

\bibitem[{Yang et~al.(2022)Yang, Song, Cho, Wang, Pan, Petzold, and
  Yu}]{yang2022oasum}
Xianjun Yang, Kaiqiang Song, Sangwoo Cho, Xiaoyang Wang, Xiaoman Pan, Linda
  Petzold, and Dong Yu. 2022.
\newblock Oasum: Large-scale open domain aspect-based summarization.
\newblock \emph{arXiv preprint arXiv:2212.09233}.

\bibitem[{Zhan et~al.(2022)Zhan, Sosea, Caragea, and Li}]{zhan2022you}
Hongli Zhan, Tiberiu Sosea, Cornelia Caragea, and Junyi~Jessy Li. 2022.
\newblock Why do you feel this way? summarizing triggers of emotions in social
  media posts.
\newblock \emph{arXiv preprint arXiv:2210.12531}.

\bibitem[{Zhang et~al.(2023)Zhang, Ladhak, Durmus, Liang, McKeown, and
  Hashimoto}]{zhang2023benchmarking}
Tianyi Zhang, Faisal Ladhak, Esin Durmus, Percy Liang, Kathleen McKeown, and
  Tatsunori~B Hashimoto. 2023.
\newblock Benchmarking large language models for news summarization.
\newblock \emph{arXiv preprint arXiv:2301.13848}.

\bibitem[{Zhong et~al.(2021)Zhong, Yin, Yu, Zaidi, Mutuma, Jha, Awadallah,
  Celikyilmaz, Liu, Qiu, and Radev}]{zhong-etal-2021-qmsum}
Ming Zhong, Da~Yin, Tao Yu, Ahmad Zaidi, Mutethia Mutuma, Rahul Jha,
  Ahmed~Hassan Awadallah, Asli Celikyilmaz, Yang Liu, Xipeng Qiu, and Dragomir
  Radev. 2021.
\newblock \href {https://doi.org/10.18653/v1/2021.naacl-main.472} {{QMS}um: A
  new benchmark for query-based multi-domain meeting summarization}.
\newblock In \emph{Proceedings of the 2021 Conference of the North American
  Chapter of the Association for Computational Linguistics: Human Language
  Technologies}, pages 5905--5921, Online. Association for Computational
  Linguistics.

\end{thebibliography}
\bibliographystyle{acl_natbib}
\clearpage
\newpage

\appendix

\onecolumn
\vspace{-20mm}
\paragraph{Appendix}
\vspace{-20mm}

\section{Generated examples}
\vspace{-2mm}
Here we show some ChatGPT-generated summaries in Table \ref{tab:case} and \ref{tab:case2}, together with their golden references.

\begin{table}[!ht]
\vspace{-2mm}
    \centering
    \footnotesize
    \resizebox{1.\linewidth}{!}{
            \begin{tabular}{llll}
            \toprule
            Zero-shot: Query in SQuALITY test set: \textit{What is the relationship like between Jeff and Ann?}\\
            \hline
            \multicolumn{4}{p{1.\linewidth}}{ 
                {\textbf{ChatGPT}:
               The relationship between Jeff and Ann is that of a married couple. They are in a restaurant discussing Jeff's job prospects when they notice a mustachioed man at the next table watching them. Ann is suspicious of the man and thinks she saw him in the parking lot. The man interrupts their conversation and offers to take them on a trip back in time. Jeff is dismissive of the idea, but Ann is intrigued and wants to know the cost. The man, Snader, gives them a card with his name and the name of his company, 4-D Travel Bureau. He invites them to take a free trial trip to a different time and promises they could be back in an hour. Jeff is skeptical, but Ann is more open to the idea. Based on the excerpt provided, it is unclear whether Jeff and Ann have any underlying issues in their relationship. They are supportive of each other and have a friendly, if somewhat exasperated, rapport. Jeff is worried about his job prospects and seems stressed, while Ann is more optimistic and tries to cheer him up. Overall, the relationship appears to be one of mutual affection and respect. }} \\
            \midrule
            \multicolumn{4}{p{1.\linewidth}}{
                {\textbf{Reference}: Jeff and Ann Elliott are a married couple. Ann is supportive of Jeff, and assures him that with their youth he will be able to rebuild his failed business. She reassures him throughout the story, even at points where it ultimately leads them into deeper trouble - such as when she tells him it wouldn’t hurt to try Mr. Snader’s time travel. Jeff is protective of Ann on several occasions, like at the start of the story suggesting he would start a brawl at the restaurant if the stranger was interested in Ann’s beauty. He is also upset enough with his business struggles that he needs to take sleeping pills, of which Ann is concerned about the amount. They remain together in the story until they are held in separate jail cells. They do not have any major disagreements in the story, and seem to enjoy their time together, only hoping to improve their lot by trying a risky time travel adventure.}}\\
            \toprule
            Zero-shot: Query in QMsum test set: \textit{summarize the discussion about the work division of the project .}\\
            \hline
            \multicolumn{4}{p{1.\linewidth}}{ 
                {\textbf{ChatGPT with full article}:
                The discussion was about the work division of the project. The project manager assigned tasks to the industrial designer, user interface designer, and marketing expert. The industrial designer was responsible for the object's appearance, while the user interface designer worked on the user experience design. The marketing expert provided insights into user preferences, which converged with the work of the user interface and industrial designers. The project manager managed the group and oversaw the project's progress. There was some confusion about the roles of the industrial and user interface designers, but it was eventually resolved.}} \\
            \midrule
            \multicolumn{4}{p{1.\linewidth}}{
                {\textbf{ChatGPT with golden article}: The project manager, Ada Longmund, is in charge of the project to create a new remote control that is original, trendy, and user-friendly. The project will have three stages of design: functional, conceptual, and detailed. The industrial designer and the user interface designer will have specific roles in the project, and the marketing expert will be in charge of the selling price of the product, which will be 25 euros. The next meeting will start in 30 minutes and the project manager will manage the whole group. The industrial designer will take care of the industrial way to transform the design and the user interface designer will design the user interface.}}\\
            \midrule
            \multicolumn{4}{p{1.\linewidth}}{
                {\textbf{Reference}: project manager would manage the whole group . industrial designer would think of the functions of the product . user interface designer would concentrate on the users ' requirement and the relation between the users and the remote control . marketing would give interface designer and industrial designer some trends about what should be done and what the users would like to have .}}\\
            \toprule
            Zero-shot: Query in QMsum test set: \textit{summarize the whole meeting .}\\
            \hline
            \multicolumn{4}{p{1.\linewidth}}{ 
                {\textbf{ChatGPT with full article}:
                The meeting was between the project manager, Ada Longmund, and the marketing, industrial design, and user interface teams. The project they are working on is to create a new and original remote control that is also user-friendly. They discussed the method of work and the specific roles of each team. The industrial designer will work on the design and transformation of the remote control, the user interface designer will work on the core functions, and the marketing expert will work on selling the product internationally. The next meeting will be in thirty minutes and each team knows what they have to do for the project. The selling price for the product will be 25 euros and the cost will not exceed 12 euros and 50 centimes. They discussed the difference between industrial design and user interface design and the project manager will manage the whole group. The meeting ended with a brief discussion on the team's knowledge of their respective jobs.}} \\
            \midrule
            \multicolumn{4}{p{1.\linewidth}}{
                {\textbf{ChatGPT with golden article}: The project manager introduces the project of creating a new remote control that is original, trendy, and user-friendly. The team will work on functional, conceptual, and detailed design with individual work and meetings. The selling price is 25 euros, and the product will be sold worldwide with a cost of no more than 12.50 euros. The team members are an industrial designer, a user interface designer, and a marketing expert. They each have a specific role to work on the design and technical functions of the remote control while thinking about user requirements. The meeting concludes with the team having thirty minutes to prepare for the next meeting.}}\\
            \midrule
            \multicolumn{4}{p{1.\linewidth}}{
                {\textbf{Reference}: the meeting , which was held at the early stage of the project , focused on the product features and work division . project manager proposed that the remote control had to be imaginative , trendy , user-friendly and international . besides , the production cost should be no more than 12.5 euros . in terms of the price , all members agreed that 25 euros would be reasonable . they also clarified each person 's duty . in the time remaining , the members did some tool training by drawing their favourite animals on the whiteboard .}}\\
            
            \bottomrule
            \end{tabular}
    }
\caption{Example of query-based summaries under the zero-shot setting. }
\label{tab:case}
\end{table}

\begin{table*}[t]
    \centering
    \footnotesize
    \resizebox{1.\linewidth}{!}{
            \begin{tabular}{llll}
            \toprule
            Zero-shot: Aspect in NEWTS test set: \textit{money, pay, paid, card, credit, paying, cards, spent, fund, amount: }\\
            \hline
            \multicolumn{4}{p{1.\linewidth}}{ 
                {
                \textbf{Article}:
              Scientists have been puzzling . for years over why some people survive Ebola while many others . perish. A new study provides strong evidence that individual . genetic differences play a major role in whether people die from . the disease. They say the discovery could help create a vaccine. Scroll down for video . Researchers think the findings may help answer somequestions about the Ebola outbreak in Guinea, Sierra Leone andLiberia, which has killed about 5,000 people. Mice used in the study were generated from eight different . strains of the animal and were bred to represent human genetic . diversity. Symptoms in these genetically diverse mice ranged . from mild weight loss to full, hemorrhagic fever, including . internal bleeding, swollen spleens and changes in liver color . and texture. 'We infected these mice with a mouse-adopted strain of Ebola . virus,' said Angela Rasmussen, a microbiologist at the . University of Washington who helped lead the study. 'In classical lab mice, this strain of Ebola kills the . animals but it doesn't produce hemorrhagic disease,' she said. Researchers at the University of Washington in Seattle . reported their findings on Thursday in the journal Science. They compared conventional lab mice, which typically just . die when infected with a mouse version of Ebola, with . genetically diverse lab mice that developed a wide range of . symptoms in much the same way that people infected with the . virus do. Mice used in the study were generated from eight different . strains of the animal and were bred to represent human genetic . diversity. Symptoms in these genetically diverse mice ranged . from mild weight loss to full, hemorrhagic fever, including . internal bleeding, swollen spleens and changes in liver color . and texture. 'We infected these mice with a mouse-adopted strain of Ebola . virus,' said Angela Rasmussen, a microbiologist at the . University of Washington who helped lead the study. 'In classical lab mice, this strain of Ebola kills the . animals but it doesn't produce hemorrhagic disease,' she said. Researchers think the findings may help answer some . questions about the Ebola outbreak in Guinea, Sierra Leone and . Liberia, which has killed about 5,000 people. A major question has been whether Ebola survivors have had . some prior viral exposure that enables their immune systems to . fight the infection or whether there is something genetically . unique about survivors that makes them resistant. The mouse study cannot address the issue of prior immunity . but makes a strong case for the idea that an individual's . genetic code plays a role in surviving Ebola. 'There is definitely a genetic component,' Rasmussen said. The study shows that host genes influence which cells become infected . The study shows that host genes influence which cells become . infected and how much the virus replicates, said Professor . Andrew Easton, a virologist at the University of Warwick, who . was not part of the study. 'Most of these genes are involved in the very earliest . stages of our immune response to infection,' something that has . also been seen with other viruses, he said in a statement. Study co-author Michael Katze of the University of . Washington said he hopes the study will quell some of the debate . over the role of genetics in disease progression. 'These mice were infected with exactly the same dose by . exactly the same route by the same investigator. 'The only thing . that was different was the genetic background,' he said. One thing that has hindered Ebola research has been the lack . of a mouse model to study the disease. Rasmussen hopes her work . now makes it easier for scientists to study drugs and vaccines . that fight Ebola. Most such studies are done in monkeys, but they can be hard . to work with. Genetically altered mice are easier to control, especially . in the strict confines of a highly secure biocontainment . laboratory where such work must be done. Rasmussen doesn't expect mice to replace monkeys, but they . may be useful for earlier-stage work.}} \\
            \midrule
            \multicolumn{4}{p{1.\linewidth}}{ 
                {
                \textbf{ChatGPT}:
               The Chancellor of the Exchequer, George Osborne, has been criticized for making the process of clawing back child benefit from better-off families too complicated. The changes, which are due to come into force in 10 weeks, will affect thousands of people earning more than £50,000 and will result in some of their pay being lost if they move in with a new partner who has children. The money will be recovered through additional tax imposed on the higher earner, regardless of whether they receive child benefit or are a parent. This will result in 500,000 people having to fill out self-assessment tax forms for the first time. The process has been described as "bizarre" and "overcomplicated" by tax experts at Deloitte. }} \\
            \midrule
            \multicolumn{4}{p{1.\linewidth}}{
                {\textbf{Reference}: Apparently, parents or parental figures in the UK receive money to help raise kids. The funds depend on their income. This may result in some financial challenges for stepparents.}}\\
            \toprule
            Zero-shot: Aspect in NEWTS test set: \textit{study, researchers, google, scientists, university, data, journal, human, humans, dna: }\\
            \hline
            \multicolumn{4}{p{1.\linewidth}}{ 
                {
                \textbf{ChatGPT}:
               A new study by researchers at the University of Washington provides strong evidence that individual genetic differences play a major role in determining whether people die from Ebola. The researchers infected genetically diverse mice with a strain of Ebola virus and found that the symptoms in the mice varied from mild weight loss to full hemorrhagic fever, similar to the symptoms seen in people infected with the virus. The study shows that host genes influence which cells become infected and how much the virus replicates. The findings may help answer some questions about the Ebola outbreak in West Africa and make it easier for scientists to study drugs and vaccines that fight Ebola. The study has been reported in the journal Science. }} \\
            \midrule
            \multicolumn{4}{p{1.\linewidth}}{
                {\textbf{Reference}: Recent studies on the fact that some people survive the Elba virus and others have not been developed by researchers, scientists. The study was developed using human data from people who survived the virus. Based on this information, tests in mice were developed in order to establish a genetic pattern resistant to the virus and thus develop an efficient vaccine. }}\\
            \toprule
            
            Zero-shot: Aspect in CovidET test set: \textit{anger}\\
            \hline
            \multicolumn{4}{p{1.\linewidth}}{ 
                {
                \textbf{Article}:
              2 weeks ago, I was feeling very confident and rsvpd yes to a company holiday party. The event is a week away and im feeling less good about it now. I really wanna skip it but i feel really guilty about it cuz i had already said yes. The thing is, my grandpa has COPD and is in the middle of a flare up. I have to be near him to help him with whatever he needs and i cant just go out and avoid him. I feel like this would be a good enough reason to skip but i live in Florida where people think covid is not real anymore. I feel like they are not gonna take me seriously and deem me as paranoid. I think taking care of my gpas health is more important than my coworkers opinion, but im still feeling quite bad at it. What would u do? Im just looking for some support and advice. I feel crazy for still caring about this virus or people make me feel crazy.}} \\
            \midrule
            \multicolumn{4}{p{1.\linewidth}}{ 
                {
                \textbf{ChatGPT}:
              I'm feeling guilty about potentially skipping a company holiday party due to my grandpa's COPD flare-up, but also worried about how my coworkers will perceive me as paranoid, and seeking support and advice on what to do.}} \\
            \midrule
            \multicolumn{4}{p{1.\linewidth}}{
                {\textbf{Reference}: I'm so tired of doubting myself because I'm worried about my coworkers' opinions.}}\\

            \bottomrule
            \end{tabular}
    }
\caption{Example of aspect-based summaries under the zero-shot setting. }
\label{tab:case2}
\end{table*}

\end{document}